\documentclass[final]{cvpr}

\usepackage{times}
\usepackage{epsfig}
\usepackage{graphicx}
\usepackage{amsmath}
\usepackage{amssymb}

\usepackage{booktabs}
\usepackage{multirow}
\usepackage{subfigure}

\usepackage{appendix}
\newcommand*{\affaddr}[1]{#1} 
\newcommand*{\affmark}[1][*]{\textsuperscript{#1}}

\usepackage[pagebackref=true,breaklinks=true,colorlinks,bookmarks=false]{hyperref}

\def\cvprPaperID{2388} 
\def\confYear{CVPR 2021}

\begin{document}

\title{Computation-Efficient Knowledge Distillation via Uncertainty-Aware Mixup}

\author{%
Guodong Xu\affmark[1], Ziwei Liu\affmark[2], Chen Change Loy\affmark[2]\\
\affaddr{\affmark[1]The Chinese University of Hong Kong}, 
\affaddr{\affmark[2]Nanyang Technological University}\\
{\tt\small xg018@ie.cuhk.edu.hk},
  {\tt\small\{ziwei.liu, ccloy\}@ntu.edu.sg}
}



\maketitle


\begin{abstract}

  Knowledge distillation, which involves extracting the ``dark knowledge'' from a teacher network to guide the learning of a student network, has emerged as an essential technique for model compression and transfer learning.
  Unlike previous works that focus on the accuracy of student network, here we study a little-explored but important question, i.e., knowledge distillation efficiency.
  Our goal is to achieve a performance comparable to conventional knowledge distillation with a lower computation cost during training.
  We show that the UNcertainty-aware mIXup (UNIX) can serve as a clean yet effective solution.
  The uncertainty sampling strategy is used to evaluate the informativeness of each training sample. 
  Adaptive mixup is applied to uncertain samples to compact knowledge.
  We further show that the redundancy of conventional knowledge distillation lies in the excessive learning of easy samples.
  By combining uncertainty and mixup, our approach reduces the redundancy and makes better use of each query to the teacher network. 
  We validate our approach on CIFAR100 and ImageNet.
  Notably, with only 79\% computation cost, we outperform conventional knowledge distillation on CIFAR100 and achieve a comparable result on ImageNet.
  The code is available at: \href{https://github.com/xuguodong03/UNIXKD}{https://github.com/xuguodong03/UNIXKD}.

\end{abstract}


\section{Introduction}

\textit{
``Young children show tremendous versatility in their learning and plasticity ... more efficient, both in how they learn and in how their brain is connected within itself.''
}

\begin{flushright}
\textit{- The Gardener and The Carpenter}
\end{flushright}

In pedagogy, learning efficiency is an important measure to evaluate the learning speed of a learner. 
A good learner not only learns well at the end but also learns fast in the process.
It is preferred if the learner can achieve a comparable or better performance with less effort.
Analogous to human learning in real world, machine learning methods~\cite{eff_survey,eff_nips} also pay attention to learning efficiency issue. The target is to accelerate training from the algorithmic aspect.

\begin{figure}[t]
	\centering
	\includegraphics[scale=0.2]{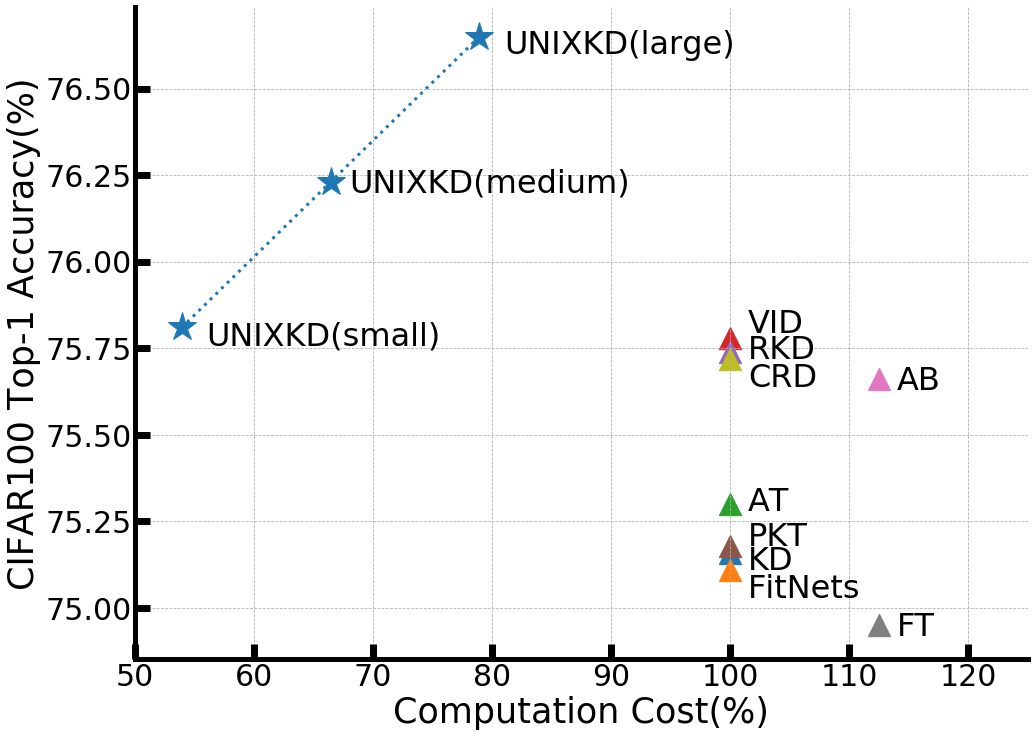}
	\caption{CIFAR100 Top-1 accuracy of different distillation methods. All the methods use the same teacher and student architectures. The computation cost of KD is set as the baseline (100\%). Hard label loss is excluded from learning objective. The cost in backbone forward/backward pass is counted and the cost of computing loss is neglected. Our method (UNIXKD) achieves comparable or better performances with a significantly lower computation cost.}
	\vspace{-5pt}
	\label{fig:teaser}
	\vspace{-5pt}
\end{figure}

Knowledge distillation (KD)~\cite{KD}, serving as a way of model compression~\cite{model_compression}, has been actively studied to deploy high-performance but cumbersome networks on edge devices. 
In KD, one typically trains a smaller network (student) under the guidance of a larger network (teacher). 
The main goal in KD is to obtain a student network with higher accuracy than that trained from scratch. 
Many studies pursue this goal by transferring better knowledge, such as intermediate features~\cite{overhaul,AB,fitnets,FSP,AT} and similarities between samples~\cite{rkd,crd,simi,sskd}. 
The student performance improves rapidly as more powerful algorithms emerge. 

Existing works pay relatively more attention to the accuracy dimension but explore little on the aspect of knowledge distillation efficiency.
The study of learning efficiency is not only practically beneficial for edge computing and budget-limited training, but also scientifically meaningful. 
For instance, it reveals the redundancy of the original method and sheds light on factors that influence the performance the most.
The study about learning efficiency in KD is scarce. In this work, we take the first step towards this little-explored but important question.

The challenge in KD efficiency lies in two aspects, \ie, how to define efficiency quantitatively and how to improve efficiency. 
Previous works in other deep learning topics usually measure efficiency from time dimension, \eg, number of iterations or epochs. 
However, different distillation methods introduce different operations into the conventional teacher-student framework, hindering a direct comparison on the cost of each iteration across the different methods.
Hence, we propose to define efficiency from the standpoint of computation cost, \ie, the number of forward/backward passes in teacher and student networks.
With this definition, different KD methods can be compared directly.
A more efficient method can achieve the same student accuracy with less computation.

Based on this definition, we propose to improve efficiency by combining two seemingly orthogonal directions, \ie, select informative samples and compact knowledge. 
The main idea is to measure the uncertainty of each sample in the student forward pass and then compact the knowledge via adaptive mixup according to their level of uncertainty.
Specifically, for the uncertain (informative) samples, a mild mixup is applied. For the certain (less informative) samples, a heavy mixup is applied. 
All the mixed images are passed as queries to teacher. 
Teacher's output logits are treated as labels to supervise the learning of the student. 

The importance of each training sample is not equal~\cite{imp_sample}. 
In the training stage, a student network has different masteries of different samples. 
The repetition of mastered samples occupies computation but brings nothing new to the network.
We adopt uncertainty to estimate the mastery of each sample. 
The occurrence frequency of samples with high confidence is reduced, so that the network can focus on those uncertain samples.
Interestingly, estimating mastery via uncertainty is closely related to active learning (AL). 
To save annotation cost, AL selects only an informative subset from the original dataset to query the oracle (expert annotators).
The KD framework is naturally similar to AL. 
The teacher plays the role of oracle and is responsible for providing labels. 
A time-consuming teacher forward pass corresponds to the label expensiveness in AL. 
Hence, we believe that the classic sampling strategy in AL, \ie, uncertainty-based sampling, is a viable strategy to decide the importance of each sample.

After obtaining the uncertainty of each sample, we compact knowledge through mixup~\cite{mixup} operation.
Mixup was originally proposed for data augmentation to improve generalization ability. 
It compacts the content in two images into a single image via pixel-wise convex combination.
Though it can compact knowledge, a disadvantage brought by mixup is that the objects in two images occlude each other, making both of them hard to recognize.
Hence, we adopt adaptive mixup according to uncertainty. A mild mixup is applied to uncertain images so that the important information in them will not be hurt by the mixing image.

\vspace{5pt}
\noindent\textbf{Contributions}.
We take the first step towards a little-explored but important question, \ie, KD efficiency.
We quantitatively define the efficiency from the standpoint of forward/backward computation cost, allowing a direct comparison across different KD methods. 
We propose a novel framework called UNIXKD to improve KD efficiency and conduct thorough experiments to validate the method.
With a significantly lower computation cost, we achieve a comparable and even better performance, compared to the conventional KD. 
We also conduct careful analyses to facilitate a deeper understanding of KD.
We show that the redundancy of KD lies in excessive learning of easy categories, and our methods can effectively reduce the redundancy.


\section{Related Work}

\noindent\textbf{Knowledge Distillation}.
The goal of KD is to transfer the knowledge from a large teacher network to a small student network. 
Hinton \etal~\cite{KD} propose to match the category distributions of the teacher and student models via KL-divergence. 
The relative probability assigned to incorrect categories encodes semantic similarity between similar categories.
This ``dark knowledge''~\cite{KD} is shown to benefit the student's learning.
Subsequent studies continue to improve the transfer of knowledge via different learning objectives. 
Some works propose to mimic the intermediate feature~\cite{fitnets} or feature's transformed variants~\cite{AB,kdsvd,FT,AINKD,FSP,AT}. 
Other works~\cite{vid,IRG,rkd,crd,sskd} extend the notion of unary matching to binary relation mimicking. 
Dabouei \etal~\cite{supermix} study the effects of data augmentation on KD and demonstrates that augmentated images can transfer extra knowledge.
All the aforementioned methods focus only on improving the accuracy of student.

A recent work by Wang \etal~\cite{activemixup} highlight the efficiency issue. They also consider both active learning and mixup to tackle the problem. 
However, our method differs significantly in the following aspects.
First, the meaning of efficiency in our work is more holistic.
Their goal is only to reduce the query numbers of teacher model, but the student cost is ignored.
In fact, if there are 20k initial unlabeled images, their method would require 2 billion student forward passes to select suitable images to query, which is not a negligible expense.
In contrast, we consider all the computation cost in the whole KD process. 
The synergy of mixup and active learning in the two works are also different.
They start from a small number of unlabeled images and use mixup to enlarge the dataset to avoid overfitting, while we use mixup to compact multiple images into one.
Finally, instead of employing active learning to reduce the billion-level image numbers into a tractable amount, we employ uncertainty to estimate the value of each original image for the subsequent adaptive mixup.

Several other methods~\cite{FSKD,FS_CVPR20,ZSKD} study the data efficiency problem in KD with the objective of obtaining high student accuracy in a data-limited setting. 
These methods typically involve synthesizing a vast number of images and using them to perform conventional KD. 
In contrast, we focus on training efficiency by conducting a holistic analysis on computation cost of student learning irrespective of the amount of labeled data.

\noindent\textbf{Active Learning}.
The idea behind active learning is that a learning agent can achieve greater accuracy with few training labels if it is allowed to select data by itself. 
An active agent poses queries in the form of unlabeled instances to be labeled by an oracle (\eg, expert annotator). 
Active learning methods work well in the setting where labels are expensive.
Since obtaining label in KD is also an expensive process, we adopt the strategy in active learning to help select informative samples. 
Common query strategies include uncertainty sampling~\cite{uncertainty,margin,entropy}, query-by-committee~\cite{qbc} and expectation-based methods~\cite{emc,eer}. We adopt uncertainty-based methods as the sampling strategy due to its low computation expense.


\section{Methodology}

The conventional knowledge distillation~\cite{KD} minimizes the distance between teacher and student softened logits through KL-divergence.
Subsequently, numerous methods explore distilling various other kinds of knowledge, such as feature map~\cite{fitnets}, attention map~\cite{AT}, pre-activation~\cite{AB} and batchnorm statistics~\cite{AINKD}.
All the distillation methods can be written in a general format:
\begin{equation}\label{eq1}
    L = d(T_t(G_t), T_s(G_s)),
\end{equation}
where $t$ and $s$ denote the teacher and student, respectively, $G_*$ denotes the feature of some network, $T_*$ denotes some pre-defined transformation, $d$ is a distance metric.
It aims at aligning the transformed teacher features $T_t(G_t)$ and transformed student features $T_s(G_s)$ in the distance space $d$.

\subsection{Computation Cost in KD}\label{sec31}

\begin{figure}[t]
	\centering
	\includegraphics[scale=0.38]{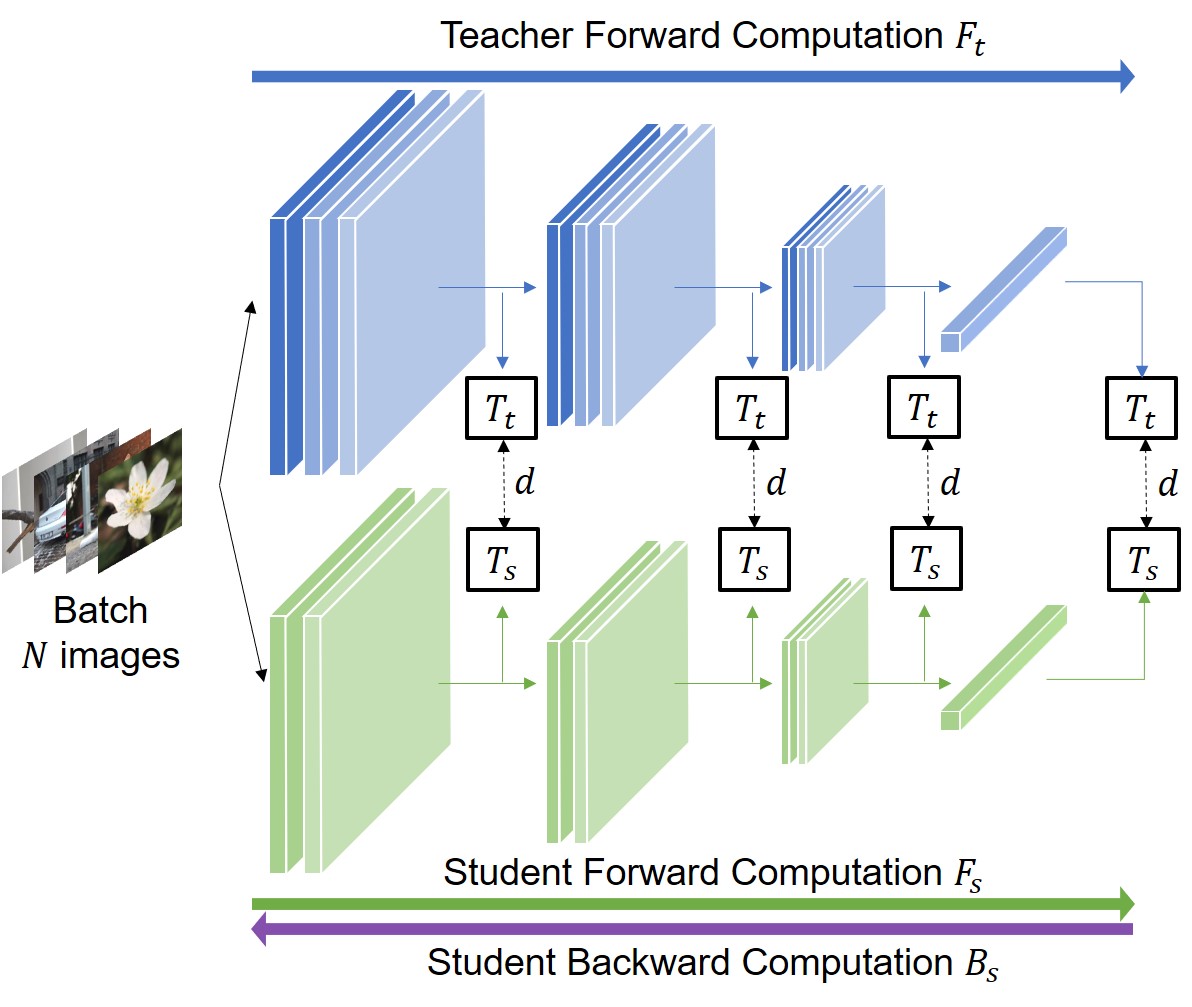}
	\caption{The general framework of distillation. The computation cost consists of three parts, \ie, teacher forward cost $F_t$, student forward cost $F_s$ and student backward cost $B_s$.}
	\label{fig:kd}
	\vspace{-15pt}
\end{figure}

\begin{figure*}[t]
	\centering
	\includegraphics[scale=0.38]{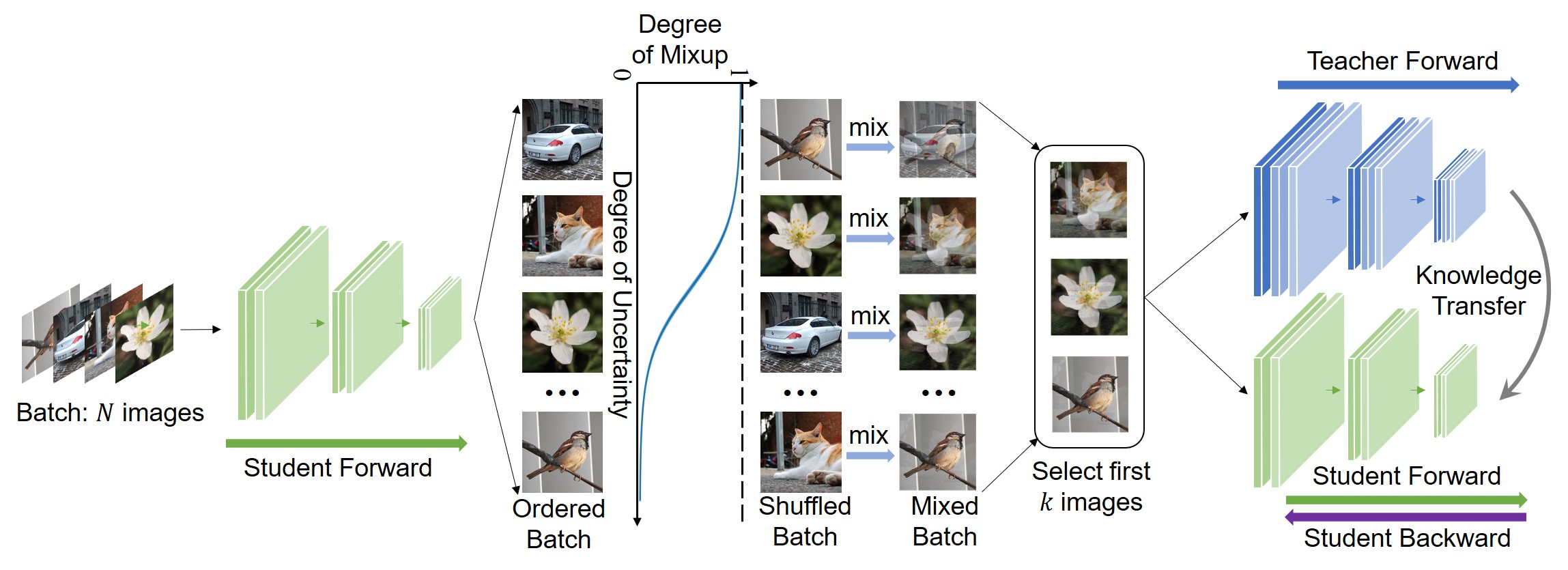}
	\caption{The framework of UNIX for knowledge distillation. Images in a batch are fed into student network and then are sorted in an uncertainty descending order. Mixup is applied adaptively according to uncertainty. For the uncertain samples, a mild mixup is applied and vice versa. Top $k$ mixed images are selected to perform a conventional KD process. Compared to conventional KD, our method saves much cost at the teacher end but increases little cost at the student end. The total cost is reduced due to the asymmetrical capacity between teacher and student.}
	\label{fig:frm}
	\vspace{-10pt}
\end{figure*}

As shown in Fig.~\ref{fig:kd}, the computation cost in KD arises from three aspects, \ie, teacher forward pass, student forward pass and student backward pass. 
Let $F_t$, $F_s$ and $B_s$ denote the floating point operation numbers in these three passes and $N$ be the batch size. 
The common practice in KD is that all the training samples would go through all the three passes.
So the total computation cost in each iteration is:
\begin{equation}\label{eq2}
    E_{kd} = N\cdot (F_t+F_s+B_s).
\end{equation}
%
A more fine-grained cost is:
\begin{equation}\label{eq3}
    E_{kd} = N_t\cdot F_t+N_{s1}\cdot F_s+N_{s2}\cdot B_s,
\end{equation}
where $N_t$, $N_{s1}$ and $N_{s2}$ are the number of samples that go through aforementioned three passes.
However, not all samples need to go through all the three passes. An efficient KD method can seek a trade-off between the three terms.
Notice that $F_t\gg F_s\approx B_s$. 
We can try to reduce the teacher query numbers $N_t$ at the expense of an increase in $N_{s1}$, but the latter is only minor therefore resulting in a net reduction in the total computation cost.  
On the contrary, previous work~\cite{activemixup} reduces $N_t$ into an acceptable range but puts no attention to the hefty increase in $N_{s1}$, resulting in a net total cost that is intractable.

\subsection{Uncertainty-Aware Mixup}\label{sec:32}

The computation of a teacher network is usually much larger than that of the student for the same input image. 
Hence, to save the total computation, we propose to reduce the number of images fed to the teacher at the expense of a slight computation increase at the student end.
Specifically, as shown in Fig.~\ref{fig:frm}, we first feed all the images in a batch to the student network and obtain their uncertainties. 
We then sort images based on their uncertainty in a descending order and select top $k$ of them to apply adaptive mixup.
Finally, the mixed images are fed into two networks to perform the KD process.
Since our method only affects the way to feed data and does not change the learning objective, it can be combined with any distillation methods.

\vspace{3pt}
\noindent\textbf{Uncertainty Estimation}.
The importance of each training sample is not equal~\cite{imp_sample}.
Considering the cost of querying teacher, we propose to feed the most informative samples to teacher network. 
We adopt the uncertainty to measure the informativeness of each training sample.
Intuitively, a sample that cannot be classified by the student confidently is a hard sample and it can bring the model with more information upon query. 
We use entropy to measure the uncertainty of classification result:
\begin{equation}\label{eq:ue}
    U_e(x) = -\sum\nolimits_{i=1}^C p_i(x) \log p_i(x),
\end{equation}
where $p_i(x)$ is the probability that sample $x$ belongs to class $i$, and $C$ is the number of class. 
A large entropy indicates a confused classification result.
Teacher's feedback of these samples would benefit the student most. 
On the contrary, samples that the student has mastered are less useful and can be discarded.

Another two common uncertainty criteria are confidence-based and margin-based measures:
\begin{equation}\label{eq:uc}
    U_c(x) = - \max_i p_i(x),
\end{equation}
\begin{equation}\label{eq:um}
    U_m(x) = -(p_i(x) - p_j(x)),
\end{equation}
where $i$ and $j$ in Eq.~\eqref{eq:um} are the first and second most probable class labels, respectively, under the model.
Confidence-based criterion considers the largest probability. 
A large $U_c$ means the model does not assign a high probability to any category, thus the sample is an uncertain sample. 
Margin-based criterion corrects the shortcoming in confidence-based one, by incorporating the posterior of the second most likely label. 
Intuitively, instances with large margins are easy, since the classifier has little doubt in differentiating between the two most probable class labels. 
We will compare the three uncertainty criteria in Sec.~\ref{sec:ablation}.

\vspace{5pt}
\noindent\textbf{Uncertainty-Aware Mixup.}
Mixup is originally a multi-sample augmentation strategy. It merges each image with another image in a pixel-wise manner:
\begin{equation}
    \lambda \sim Beta(\alpha,\alpha),
\end{equation}
\begin{equation}
    x = \lambda x_i + (1-\lambda)x_j,
\end{equation}
where $x_i$ and $x_j$ are the $i$-th and $j$-th image, respectively, and a merging coefficient $\lambda$ is sampled from a Beta distribution parameterized by $\alpha$.
In original Mixup~\cite{mixup}, the $x$'s label is a convexed combination of $x_i$'s label and $x_j$'s label. 
However, in this work, we feed the mixed images to teacher network and treat teacher's logits as the groundtruth label. 

The goal of mixup in our work is not to augment the original data, but to compress the content in two images into a single image. 
We hope that a mixed image can transfer more knowledge than a normal image. 
However, compression usually brings information loss. 
A pixel-wise merging leads to mutual aliasing between two images.
The visual pattern in each image is hurt by the mixing image, making the synthesized image blurry and semantically meaningless.

\begin{table*}[]
    \centering
    \caption{Results on cross-architecture pairs: CIFAR100 Top-1 accuracy and computation cost. \textbf{Bold} font denotes the result that outperforms KD. Our method averagely surpasses KD by 1.07\% with 79.44\% computation cost. Average on 4 runs.}
    \vspace{5pt}
    \small{
    \begin{tabular}{cc|cccccc}
        \toprule
        \multicolumn{2}{c|}{Teacher} & VGG13       & ResNet50      & ResNet50  & resnet32$\times$4 & resnet32$\times$4 & WRN-40-2  \\
        \multicolumn{2}{c|}{Student} & MobileNetV2 & MobileNetV2   & VGG8      & ShuffleV1         & ShuffleV2         & ShuffleV1 \\
        \midrule
        \multicolumn{2}{c|}{$F_t / F_s$} & 38.17 & 174.00 & 13.56 & 27.16 & 23.49 & 8.22\\
        
        \midrule
        \multirow{2}*{KD} & Acc & 68.26 & 68.26 & 73.46 & 74.45 & 75.16 & 75.78 \\ 
        ~ & Computation & 100\% & 100\% & 100\% & 100\% & 100\% & 100\%\\
        
        \midrule
        \multirow{2}*{\shortstack{Random+KD\\$k$=48}} & Acc & 66.99 & 65.72 & 73.32 & 73.69 & 74.68 & 75.70 \\ 
        ~ & Computation & 75\% & 75\% & 75\% & 75\% & 75\% & 75\%\\
        
        \midrule
        \multirow{2}{*}{\shortstack{UNIXKD\\$k$=40}} & Acc & 67.46 & 67.78 & \textbf{73.62} & \textbf{75.44} & \textbf{76.23} & \textbf{76.59}\\
        ~ & Computation & 64.99\% & 63.07\% & 68.93\% & 65.93\% & 66.42\% & 72.29\%\\
        
        \midrule
        \multirow{2}{*}{\shortstack{UNIXKD\\$k$=48}} & Acc & \textbf{68.47} & \textbf{69.06} & \textbf{74.24} & \textbf{76.41} & \textbf{76.65} & \textbf{76.92}\\
        ~ & Computation & 77.49\% & 75.57\% & 81.43\% & 78.43\% & 78.92\% & 84.79\%\\
        
        \bottomrule
    \end{tabular}
    }
    \label{tab:diff_arch}
    \vspace{-10pt}
\end{table*}

To make use of the compression effect while reducing the destruction to informative samples, we apply mixup operation in an adaptive manner.
Specifically, we sort the samples in a batch in a descending order according to their uncertainties (Fig.~\ref{fig:frm}).
We also shuffle the original batch to prepare the mixing data.
Before the mixup between sorted version and shuffled version, we introduce a correction factor $c$ to control the mixup level:
\begin{equation}
    x = (1-c\cdot\lambda)x_{\mathrm{sort}} + c\cdot\lambda~x_{\mathrm{shuffle}},
\end{equation}
where $x_{\mathrm{sort}}$ and $x_{\mathrm{shuffle}}$ are both 4D tensors with the first dimension equals to the batchsize, the correction factor $c$ is a real number that belongs to interval $[0,1]$.
The $c$'s value is related to the uncertainty ranking of each sample. 
For an uncertain sample, $c$ is set to a small number and $1-c\cdot\lambda$ is close to 1, so that mixup is mild and vice versa.
In fact, $c$ can be any monotonic increasing function of uncertainty ranking.
In this work, we select the sigmoid function:
\begin{equation}
    c = \mathrm{sigmoid}(w\cdot\frac{ranking-b}{batchsize}),
\end{equation}
where the $b$ and $w$ are parameters to control the position and shape of sigmoid function. 
As $w$ increases from small to large, the sigmoid function changes from a linear function to a step function.
Finally, we select the top $k$ images from all the mixed images and feed them into the teacher and student networks. The distillation loss in Eq.~\eqref{eq1} is computed on these mixed images.

\vspace{5pt}
\noindent\textbf{Computation Cost Analysis}.
The computation in our method comes from two parts, \ie, uncertainty estimation and distillation of mixed images. The total computation cost in each iteration is:
\begin{equation}
    \begin{aligned}
    \label{eq:finalcost}
    E &= N\cdot F_s + k\cdot(F_t+F_s+B_s)\\
    & = k\cdot F_t + (N+k)\cdot F_s + k\cdot B_s
    \end{aligned}
\end{equation}
Compared to conventional KD in Eq.~\eqref{eq2}, our method increases the number of student forward passes from $N$ to $N+k$ and reduces the teacher forward passes from $N$ to $k$. 
Considering $F_t$ is usually one order larger than $F_s$, the total cost is reduced.


\section{Experiments}

\begin{table*}[]
    \centering
    \caption{Results on similar-architecture pairs: CIFAR100 Top-1 accuracy and computation cost. \textbf{Bold} font denotes the result that outperforms KD. Our method averagely surpasses KD by 0.49\% with 92.40\% computation cost. The marginal superiority is due to the small gap between teacher and student computation cost. Average on 4 runs.}
    \vspace{5pt}
    \small{
    \begin{tabular}{cc|cccccc}
        \toprule
        \multicolumn{2}{c|}{Teacher} & WRN-40-2 & WRN-40-2 & resnet56 & resnet32$\times$4 & VGG13 & VGG16\\
        \multicolumn{2}{c|}{Student} & WRN-16-2 & WRN-40-1 & resnet20 & resnet8$\times$4 & VGG8 & VGG8\\
        \midrule
        \multicolumn{2}{c|}{$F_t / F_s$} & 3.25 & 3.93 & 3.06 & 6.12 & 2.97 & 4.14\\
        
        \midrule
        \multirow{2}*{KD} & Acc & 75.06 & 73.95 & 70.94 & 73.54 & 73.50 & 72.34\\ 
        ~ & Computation & 100\% & 100\% & 100\% & 100\% & 100\% & 100\%\\
        
        \midrule
        \multirow{2}*{\shortstack{Random+KD\\$k$=48}} & Acc & 74.73 & 73.55 & 70.09 & 72.87 & 72.62 & 71.68 \\ 
        ~ & Computation & 75\% & 75\% & 75\% & 75\% & 75\% & 75\%\\
        
        \midrule
        \multirow{2}{*}{\shortstack{UNIXKD\\$k$=36}} & Acc & \textbf{75.19} & 73.51 & 70.06 & \textbf{74.26} & 73.18 & \textbf{72.38} \\
        ~ & Computation & 75.31\% & 73.13\%	& 76.01\% & 68.57\% & 76.35\% & 72.53\%\\
        
        \midrule
        \multirow{2}{*}{\shortstack{UNIXKD\\$k$=48}} & Acc & \textbf{75.69} & \textbf{74.72} & 70.77 & \textbf{74.67} & \textbf{73.56} & \textbf{72.77} \\
        ~ & Computation & 94.06\% & 91.88\% & 94.76\% & 87.32\% & 95.10\% & 91.29\%\\
        
        \bottomrule
    \end{tabular}
    }
    \label{tab:same_arch}
\end{table*}

\begin{table*}[]
    \centering
    \caption{CIFAR100 Top-1 accuracy and computation cost when applying UNIX to AT~\cite{AT} and SP~\cite{simi}. \textbf{Bold} font denotes the result that outperforms original distillation method. On both similar and different architecture settings, UNIX improves the efficiency of AT and SP.}
    \vspace{5pt}
    \small{
    \begin{tabular}{ccc|ccc|ccc}
        \toprule
        \multirow{2}*{Teacher} & \multirow{2}*{Student} & \multirow{2}*{$F_t/F_s$} & 
         \multirow{2}*{KD+AT} & \multirow{2}*{\shortstack{Random\\+KD+AT}} & \multirow{2}*{\shortstack{UNIXKD+AT}} & \multirow{2}*{KD+SP} & \multirow{2}*{\shortstack{Random\\+KD+SP}}  & \multirow{2}*{\shortstack{UNIXKD+SP}} \\
         ~ & ~ & ~ & ~ & ~ & ~ & ~ & ~ & ~ \\
        \midrule
        \multirow{2}*{WRN-40-2} & \multirow{2}*{WRN-16-2} & \multirow{2}*{3.25} & 75.16  & 74.85 & \textbf{75.29}  & 74.64  & 73.67  & \textbf{74.66} \\
        ~ & ~ & ~ & 100\% & 75\% & 75.31\% & 100\% & 75\% & 75.31\% \\
        \midrule
        \multirow{2}*{resnet32$\times$4} & \multirow{2}*{ShuffleV2} & \multirow{2}*{23.49} & 75.70 & 74.73 & \textbf{76.50} & 75.74 & 75.05 & \textbf{77.12} \\
        ~ & ~ & ~ & 100\% & 78.13\% & 78.92\% & 100\% & 78.13\% & 78.92\% \\
        \bottomrule
    \end{tabular}
    }
    \label{tab:other_kd}
    \vspace{-8pt}
\end{table*}

\begin{table}[]
    \centering
    \caption{ImageNet Top-$k$ accuracy and computation cost. UNIXKD outperforms KD in label-based setting and achieves comparable results in label-free setting. We use a batchsize of 256. We set $k$=200 for Random and $k$=192 for UNIXKD.}
    \vspace{5pt}
    \small{
    \begin{tabular}{c|ccc|cc}
        \toprule
        ~ & \multirow{2}*{KD} & \multirow{2}*{\shortstack{Random\\+KD}} & \multirow{2}*{UNIXKD} & \multirow{2}*{\shortstack{KD\\+label}} & \multirow{2}*{\shortstack{UNIXKD\\+label}}\\
        ~ & ~ & ~ & ~ & ~ & ~ \\
        \midrule
        Top-1 & 45.16 & 44.28 & 45.11 & 53.53 & 53.64\\
        Top-5 & 72.57 & 71.79 & 72.73 & 77.75 & 78.39\\
        \midrule
        Comp. & 100\% & 78.13\% & 77.23\% & 100\% & 77.23\%\\
        \bottomrule
    \end{tabular}}\label{tab:imagenet}
    \vspace{-10pt}
\end{table}

The experiments section consists of three parts. In Sec.~\ref{sec:comp}, we apply UNIX in conventional KD and demonstrate its ability to save computation. Ablation studies are conducted in Sec.~\ref{sec:ablation} to examine the effectiveness of each designed component. In Sec.~\ref{sec:ana}, we conduct thorough analyses to understand how our method works.

Evaluations are conducted on CIFAR100~\cite{cifar100} and ImageNet~\cite{imagenet} datasets, both of which are widely used as the benchmarks for KD.
CIFAR100 consists of $60,000$ $32\times32$ colour images, including $50,000$ images for training and $10,000$ images for testing. There are 100 classes, each contains 600 images. 
ImageNet is a large-scale classification dataset, containing $1,281,167$ images for training and $50,000$ images for testing.

Since there is no method designed for improving efficiency in KD~\footnote{Although Wang~\etal~\cite{activemixup} involve efficiency, they require the dataset to be a small image pool instead of the whole dataset in conventional KD, making the comparison with our method not suitable.}, we compare our method with two baselines, \ie, conventional KD and Random.
Our goal is to achieve a comparable performance with conventional KD with a smaller computation cost. 
Hence, we regard KD as the upper bound.
To demonstrate the improvement brought by our method, we select the Random, \ie, randomly select $k$ images from a batch to perform KD, as the lower bound.
We set batchsize as 64 and 256 in CIFAR100 and ImageNet experiments, respectively.
We select VGG~\cite{vgg}, ResNet~\cite{resnet}, WRN~\cite{wrn}, ShuffleNet~\cite{shufflenetv2,shufflenet} and MobileNetV2~\cite{mobilenet} as teacher and student networks. 

\subsection{Comparative Results}\label{sec:comp}

\noindent\textbf{CIFAR100}.
We compare performances on 11 teacher-student pairs to eliminate architecture influence. 
We split them into two groups according to whether teacher and student have similar architecture styles. 
The results are shown in Table~\ref{tab:diff_arch} and Table~\ref{tab:same_arch}. 
In each table, we list $F_t/F_s$, the computation cost ratio, to facilitate understanding about the capacity gap between teacher and student. For all the methods, we list their accuracy and computation. We regard KD's computation as the baseline ($100\%$). 
To study the efficiency in the transferring process, we \textit{exclude the hard label loss} in conventional KD for all the methods. Our method is still advantageous after adding the hard label loss. The results are provided in the Sec.~\ref{sec:a-cifar}.

For teacher-student pairs with different architectures (Table~\ref{tab:diff_arch}), our method ($k=40$, the number of top images from all the mixed images, see Eq.~\eqref{eq:finalcost}) outperforms KD on four out of six pairs. 
On average, it surpasses KD by $0.29\%$ with $66.94\%$ computation cost. 
When we increase $k$ from $40$ to $48$, our method outperforms KD on all the pairs. 
On average, it surpasses KD by $1.07\%$ with $79.44\%$ computation cost.
With a similar architecture setting (Table~\ref{tab:same_arch}), our method ($k=36$) slightly fall behind KD by $0.11\%$ accuracy with 73.65\% computation. 
By increasing $k$ to $48$, it outperforms KD by 0.49\% with 92.40\% computation. . 
The difference in the degree of improvement in the two settings lies in the difference in $F_t/F_s$.
In Table~\ref{tab:diff_arch}, the student networks are mostly lightweight models. 
The computation is dominated by the number of teacher forward passes. 
However, in Table~\ref{tab:same_arch}, the computation of teacher and student is roughly of the same magnitude.
Therefore, the decrease of teacher forward pass is compensated by the increase at the student end, causing a marginal difference in the total computation.
Hence, our method is more applicable in the setting where the computation gap between teacher and student is large.

\begin{table*}[]
    \centering
    \caption{Ablation study on CIFAR100 dataset. We compare four variants of UNIXKD with KD and Random. We list their mean accuracies on six teacher-student pairs. \textbf{Bold} font denotes result that outperforms KD. The two main components in our methods, \ie, uncertainty and mixup, can both achieve comparable performances with KD. The combination leads to further improvements. Average on 4 runs. }
    \vspace{5pt}
    \small{
    \begin{tabular}{cc|cccccc|c}
        \toprule
        \multicolumn{2}{c|}{Teacher} & WRN40-2    & resnet32$\times$4 & ResNet50  & resnet32$\times$4 & resnet32$\times$4 & WRN40-2 & \multirow{2}*{Mean} \\
        \multicolumn{2}{c|}{Student} & WRN16-2    & resnet8$\times$4  & VGG8      & ShuffleV1         & ShuffleV2         & ShuffleV1 & ~\\
        
        \midrule
        \multicolumn{2}{c|}{KD} & 75.06 & 73.54 & 73.46 & 74.45 & 75.16 & 75.78 & 74.58\\ 
        
        \midrule
        \multicolumn{2}{c|}{Random $k$=48} & 74.73 & 72.87 & 73.32 & 73.69 & 74.68 & 75.70 & 74.17\\ 
        
        \midrule
        \multirow{3}*{\shortstack{Uncertainty\\(w/o Mixup)}} & 
        Confidence & \textbf{75.34} & \textbf{73.79} & 73.30 & 73.88 & 74.93 & 75.39 & 74.44\\ 
        ~ & Margin & \textbf{75.50} & \textbf{74.09} & 73.16 & 73.97 & \textbf{75.20} & 75.43 & 74.56\\
        ~ & Entropy & \textbf{75.24} & \textbf{74.17} & 73.27 & 74.27 & 75.01 & \textbf{75.86} & \textbf{74.64}\\
        
        \midrule
        \multicolumn{2}{c|}{Mixup w/o Uncertainty} & 73.59 & 72.69 & 73.11 & \textbf{75.93} & \textbf{75.72} & \textbf{76.36} & 74.57\\
        
        \midrule
        \multicolumn{2}{c|}{Non-adaptive Mixup} & 74.29 & 73.34 & \textbf{73.63} & \textbf{76.03} & \textbf{76.03} & \textbf{76.12} & \textbf{74.91}\\
        
        \midrule
        \multirow{3}*{UNIXKD} & 
        $w=1$ & \textbf{75.69} & \textbf{74.46} & \textbf{73.97} & \textbf{75.80} & \textbf{76.23} & \textbf{77.09} & \textbf{75.54}\\ 
        ~ & $w=10$ & \textbf{75.59} & \textbf{74.67} & \textbf{74.24} & \textbf{76.41} & \textbf{76.65} & \textbf{76.92} & \textbf{75.75}\\
        ~ & $w=1000$ & 74.93 & \textbf{73.96} & \textbf{73.86} & \textbf{76.32} & \textbf{76.47} & \textbf{76.16} & \textbf{75.28}\\
        
        \bottomrule
    \end{tabular}}\label{tab:ablation}
    \vspace{-10pt}
\end{table*}

\vspace{5pt}
\noindent\textbf{Combine with Other KD Methods}.
Besides conventional KD~\cite{KD}, we also combine UNIX with other distillation methods.
Specifically, we select AT~\cite{AT} and SP~\cite{simi}, for they represent two mainstream methods in distillation, \ie, mimicking intermediate features and mimicking relations between samples, respectively.
Since the learning objective in the two methods cannot update the classifier layer, we combine them with conventional KD, namely the final loss is the combination of KL divergence and respective mimicking loss.
We also select two teacher-student pairs to examine our approach under similar and different architecture settings.
As shown in Table~\ref{tab:other_kd}, UNIXKD outperforms AT and SP on both similar-architecture and cross-architecture settings, demonstrating the applicability of UNIX to various distillation methods.

\vspace{5pt}
\noindent\textbf{ImageNet}.
Limited by computation resource, we only conduct one teacher-student pair on ImageNet, \ie, ResNet18 as teacher and ShuffleV2$\times$0.5 as student. 
As shown in Table~\ref{tab:imagenet}, with just 77.23\% computation cost, UNIXKD outperforms KD on both Top-1 and Top-5 accuracies in label-based setting. It also achieves comparable results in label-free setting.
The results on ImageNet demonstrate the scalability of UNIXKD to large-scale dataset.

\subsection{Ablation Study}\label{sec:ablation}

We selectively disable certain parts of the whole framework to examine their effects. 
The experiments are conducted on CIFAR100.
Specifically, there are four variants: 
    1) Uncertainty without Mixup, namely select the top-$k$ uncertain samples and feed them to networks directly, 
    2) Mixup without Uncertainty, \ie, randomly select $k$ images and apply conventional mixup, 
    3) Non-adaptive Mixup, namely select top-$k$ uncertain samples and apply conventional mixup,
    4) UNIXKD, \ie, the full version of our method. 
For all the variants, we take $k$=48.
We list the mean accuracy averaged on six teacher-student pairs to show an overall performance.
The results are shown in Table~\ref{tab:ablation}.

\vspace{3pt}
\noindent\textbf{Effect of Uncertainty Strategy.}
We eliminate the influence of mixup and examine the improvement brought by uncertainty sampling. 
As discussed in Sec.~\ref{sec:32}, we compare three sampling strategies, \ie, Confidence, Margin and Entropy.
As shown in Table~\ref{tab:ablation}, all three uncertainty strategies outperform Random and achieve comparable results with KD. 
Notably, without the help of mixup, entropy alone can surpass KD.
Among these strategies, confidence behaves worse than the other two methods, because it only takes the maximum probability into account and ignores the full distribution in logits.

\begin{figure}[ht]
    \subfigure[Distance to category center]
    {\label{fig:property1}\includegraphics[scale=0.19]{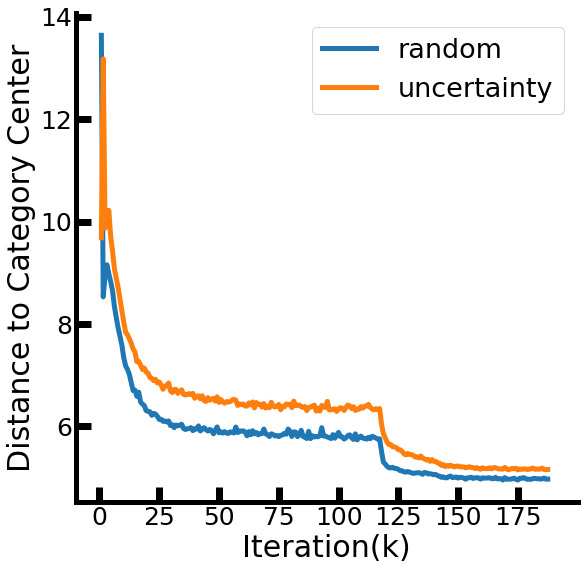}}
    \subfigure[Teacher entropy]
    {\label{fig:property2}\includegraphics[scale=0.19]{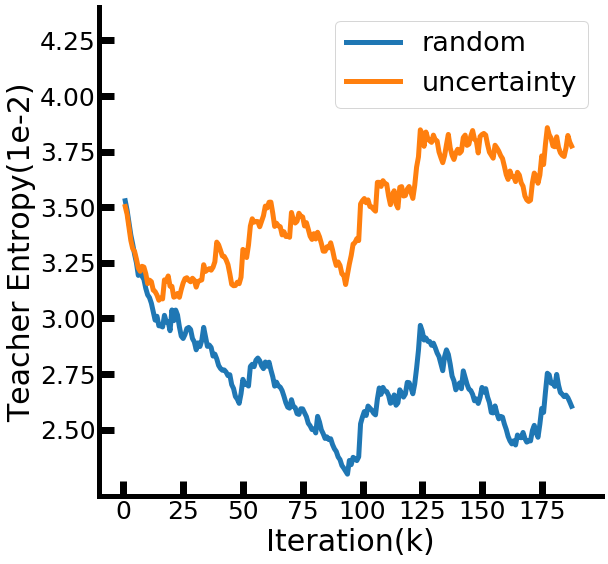}}
    \caption{The property of selected samples. Uncertainty strategy prefers to select samples near the decision boundary. These samples are regarded as hard by both teacher and student.}
    \vspace{-10pt}
\end{figure}

\vspace{3pt}
\noindent\textbf{Effect of Mixup.}
We investigate Mixup's effect by removing the uncertainty estimation. 
As shown in Table~\ref{tab:ablation}, it achieves a similar mean accuracy compared to KD.
However, its variance among different pairs is large. 
It outperforms KD by a large margin on the last three pairs while it falls behind KD on the first three pairs.
We conjecture that mixup's occasional accuracy degradation is a result of the aliasing effect on informative samples. 
To examine it, we use Uncertainty and Mixup simultaneously while combining them in a non-adaptive manner.
The accuracy gap between this variant and the full version shows that adaptive mixup is necessary for mitigating the aliasing effect.

\vspace{3pt}
\noindent\textbf{Effect of Correction Factor.}
As discussed in Sec.~\ref{sec:32}, the degree of adaptive mixup is controlled by the correction factor $c$,
which is the function of uncertainty ranking, and its shape is controlled by the parameter $w$. 
We study three cases:
    1) $w$=1, $c$ is a linear function of uncertainty ranking,
    2) $w$=10, $c$ has a typical sigmoid shape,
    3) $w$=1000, $c$ becomes a step function; it is equivalent to applying no mixup to uncertainty samples and applying conventional mixup to certain samples.
As shown in Table~\ref{tab:ablation} (bottom), the mean accuracy shows a rise-and-fall trend as $w$ increases from $1$ to $1000$. 
And $w$=10 achieves the best results.

\begin{figure}[t]
    \subfigure[Effect of number of sampling]
    {\label{fig:NvsACC}\includegraphics[scale=0.19]{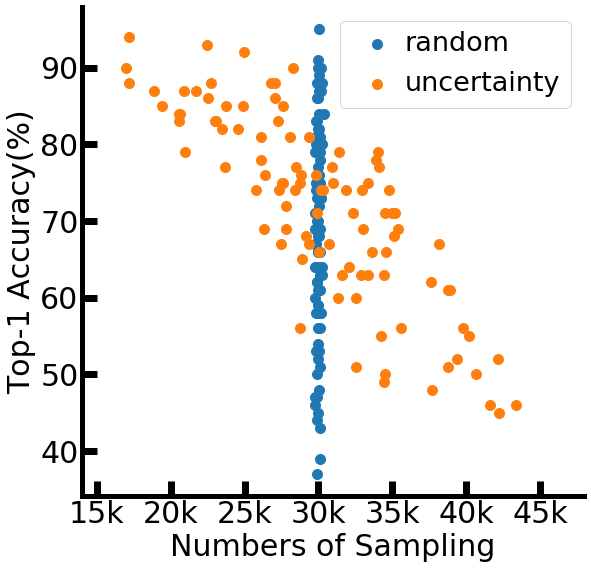}}
    \subfigure[Category accuracy]
    {\label{fig:CvsACC}\includegraphics[scale=0.19]{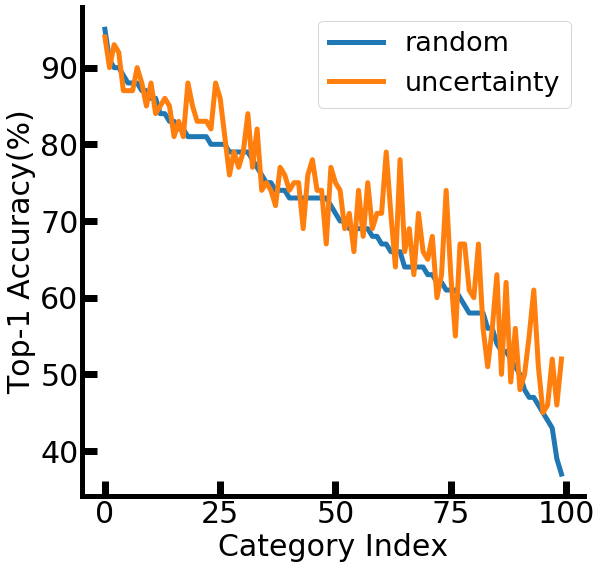}}
    \caption{Uncertainty strategy selects more samples on hard classes and less samples on easy classes. It greatly improves the performances on hard classes while maintaining comparable performances on easy classes, making the total performance increase.}
    \vspace{-10pt}
\end{figure}

\subsection{Further Analysis}\label{sec:ana}

We further analyze the uncertainty criterion to help understand how it improves accuracy and efficiency.
All the experiments are conducted on CIFAR100.

\vspace{5pt}
\noindent\textbf{The Property of Selected Samples.}
We study the sample distribution in feature space by computing the distance between each selected sample and its corresponding category center.
We use resnet32$\times$4 and ShuffleV2 as teacher and student networks, respectively. 
As shown in Fig.~\ref{fig:property1}, uncertainty strategy prefers to select samples far away from the category center, \ie, samples near the decision boundary. 
These samples are more informative than centered samples for the student to distinguish different categories.
We also compute teacher's entropy of selected samples as illustrated in Fig.~\ref{fig:property2}.
For the hard samples selected by student, teacher also regards them as hard, demonstrating that the sample difficulty is an inherent property of data, and an uncertainty strategy can discover informative samples from a nondistinctive dataset.
Besides, teacher's entropy increases as number of iteration increases, showing that Uncertainty's selection ability evolves as training proceeds.

\vspace{5pt}
\noindent\textbf{How Uncertainty Sampling Improves Accuracy?}
Considering UNIX changes the occurrence frequency of each sample in the whole training stage, we study the relation between the sampling numbers of each category and the corresponding category accuracy.

We use resnet32$\times$4 and resnet8$\times$4 as teacher and student, respectively. 
The result is shown in Fig.~\ref{fig:NvsACC}. Each node in the figure represents a single category. 
The $x$-axis denotes the total times that each category has been sampled in the training process. 
The $y$-axis denotes the accuracy of each category. 
For Random baseline, the sampling times of different categories are roughly the same.
However, for uncertainty strategy, there is an obvious negative correlation.
Uncertainty tends to select more samples on hard categories and fewer samples on easy categories. 
To have a better view of two methods' different behaviors on different categories, we arrange the category index in the descending order of accuracy yielded by Random.
As shown in Fig.~\ref{fig:CvsACC}, Uncertainty achieves higher accuracy on hard categories, benefited from their large sampling numbers. 
Though uncertainty strategy samples less on easy categories, its accuracy on easy categories remains comparable to that of Random.
Hence, the overall accuracy of uncertainty is better than Random.
The result suggests that the excessive learning of easy categories causes redundancy in conventional KD.
Our method reduces this redundancy by reasonably adjusting the occurrence frequencies of different categories, thus achieving a comparable result with lower computation.

\begin{figure}[t]
    \subfigure[Random sampling pattern]
    {\label{fig:heatmap_random}\includegraphics[scale=0.19]{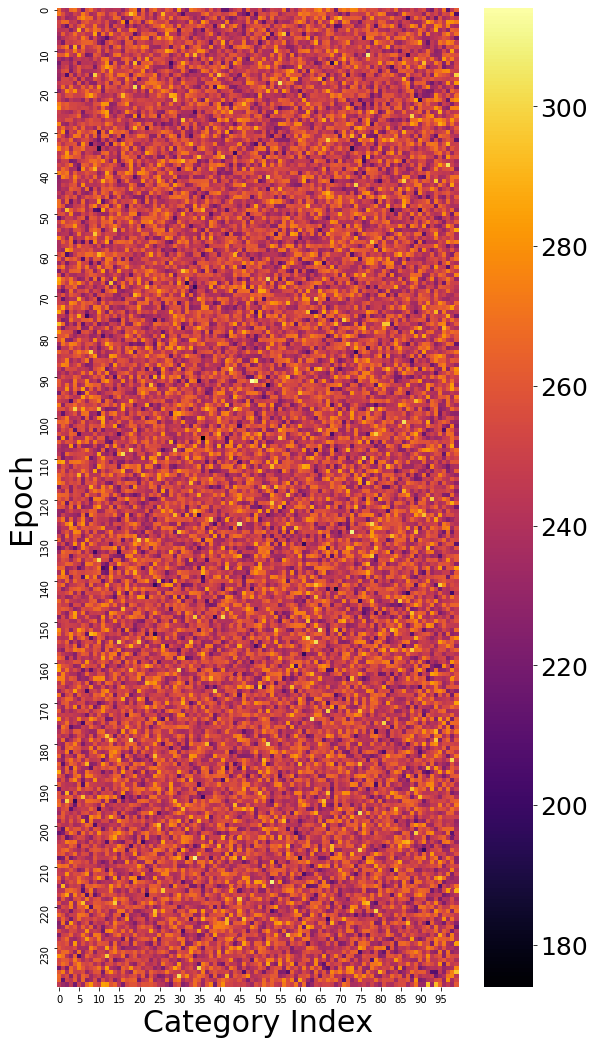}}
    \subfigure[Uncertainty sampling pattern]
    {\label{fig:heatmap_uncertainty}\includegraphics[scale=0.19]{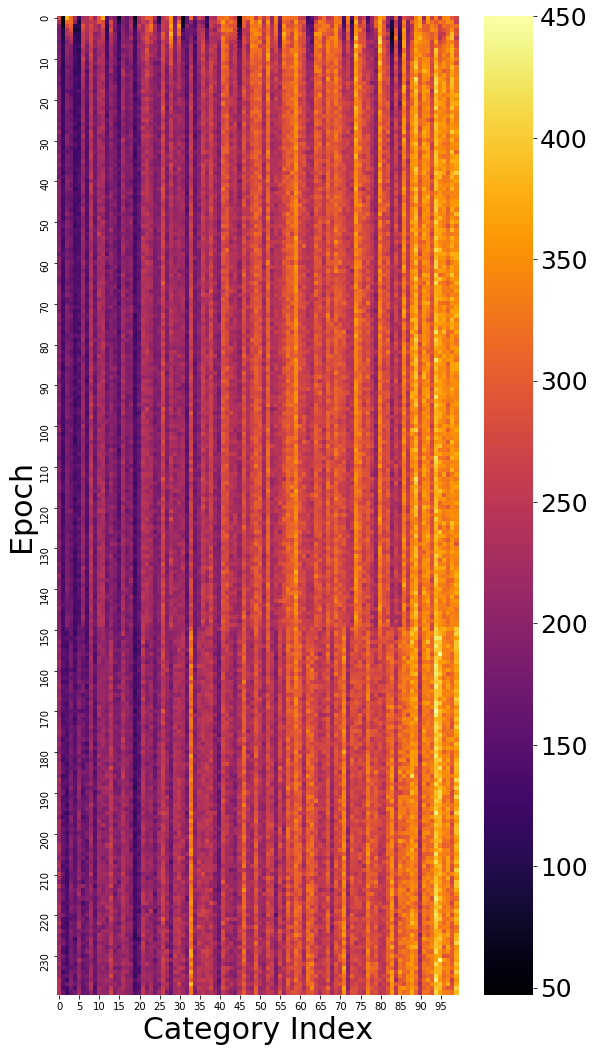}}
    \caption{Random sampling pattern varies little as training proceeds. Uncertainty learns hard category sampling pattern rapidly in the first several epochs, and keeps stable in following epochs.}
    \vspace{-10pt}
    \label{fig:heatmap}
\end{figure}

\vspace{5pt}
\noindent\textbf{How Sampling Changes with Epoches?}
To understand student's learning behavior along the time dimension, we study how quick the uncertainty criterion is in picking the hard sampling strategy.
Specifically, we count the sample numbers of different categories in each epoch and illustrate them through heatmaps in Fig.~\ref{fig:heatmap}. 
The category index along $x$-axis is arranged in a descending order according to the category accuracy of final epoch.
Uncertainty's trend along the category dimension corresponds to hard category sampling. 
From the time dimension, most of the changes happen at the first several epochs.
It indicates that our method learns hard category sampling strategy in a fast speed and then maintains a stable sampling pattern until the end of the training. 
As a comparison, Random strategy's pattern varies little along both the time and category dimensions.


\section{Conclusion}

In this work, we pay attention to a little-explored but important question in knowledge distillation, \ie, KD efficiency. 
The objective is to reduce the computation cost in KD without compromising its performance.
We proposed a novel framework called UNIX to tackle this question.
It evaluates the informativeness of each training sample via uncertainty estimation and then applies an adaptive mixup to compact more knowledge into a single sample.
We validate our method by conducting thorough experiments on CIFAR100 and ImageNet.
Our method achieves better result than conventional KD with less computation, demonstrating the effectiveness of our approach.
Further analyses show that our method can effectively reduce the redundancy in knowledge distillation.

{\small
\bibliographystyle{ieee_fullname}
\bibliography{egbib}
}

\clearpage
\begin{appendix}
In appendix, we demonstrate the label-free results (training without hard label loss) on TinyImageNet~\cite{tinyimagenet} in Sec.~\ref{sec:a-tiny} and the label-based (training with hard label loss) results on CIFAR100~\cite{cifar100} in Sec.~\ref{sec:a-cifar}. In Sec.~\ref{sec:a-implementation}, we provide implementation details of UNIXKD, including network architectures, data augmentation and training hyperparameters.

\section{Results on TinyImageNet}\label{sec:a-tiny}

We select two teacher-student pairs in TinyImageNet experiments.
For all the comparing methods, we exclude the hard label loss in learning objective.
As shown in Table~\ref{tab:a-tiny}, UNIXKD outperforms KD by using fewer than 79\% of the orginal computation cost.

\section{Label-Based Results on CIFAR100}\label{sec:a-cifar}

To study the efficiency in transferring process, we exclude the hard label loss in for the experiments in main paper. 
We also conduct experiments where hard label loss is incorporated into the learning objective. 
The results are shown in Table~\ref{tab:a-diff_arch} and Table~\ref{tab:a-same_arch}. 
On cross-architecture pairs, our method averagely outperforms KD+label by 1.10\% with 79.44\% computation cost.
On similar-achitecture pairs, our methods averagely outperforms KD+label by 0.53\% with 92.40\% computation cost.
The result shows UNIXKD's adaptation ability in various settings.

\section{Implementation Details}\label{sec:a-implementation}

\subsection{Network Architectures}

On CIFAR100, we adopt resnet~\cite{resnet}, ResNet~\cite{resnet}, WideResNet~\cite{wrn}, MobileNet~\cite{mobilenet}, VGG~\cite{vgg} and ShuffleNet~\cite{shufflenetv2,shufflenet} as the network backbones.
For resnet, we use resnet-d to represent CIFAR-style resnet with three groups of basic blocks, each with 16, 32 and 64 channels, respectively. 
resnet8$\times$4 and resnet32$\times$4 indicate a $4\times$ wider network.
For ResNet, ResNet-d represents ImageNet-style ResNet with Bottleneck blocks and more channels.
For WideResNet (WRN), WRN-d-w represents wide ResNet with depth $d$ and width factor $w$. 
For MobileNet, following ~\cite{crd}, we use a width multiplier of 0.5.
For vgg, ShuffleNetV1 and ShuffleNetV2, we adapt their architectures to CIFAR100~\cite{cifar100} dataset from their original ImageNet~\cite{imagenet} counterparts.
On ImageNet, we use the PyTorch~\cite{pytorch} official implementation of ResNet18~\cite{resnet} and ShuffleV2$\times$0.5~\cite{shufflenetv2}.

\begin{table}[t]
    \centering
    \caption{TinyImageNet Top-1 accuracy and computation cost of different methods. \textbf{Bold} font denotes the result that outperforms KD. We use a batchsize of 256 for all the methods.}
    \vspace{5pt}
    \small{
    \begin{tabular}{cc|ccc}
        \toprule
        \multicolumn{2}{c|}{Teacher} & resnet32$\times$4   & resnet32$\times$4 \\
        \multicolumn{2}{c|}{Student} & ShuffleV1           & ShuffleV2         \\
        \midrule
        \multicolumn{2}{c|}{$F_t / F_s$} & 27.16 & 23.49\\
        
        \midrule
        \multirow{2}*{KD} & Acc & 62.16 & 62.83 \\ 
        ~ & Computation & 100\% & 100\% \\
        
        \midrule
        \multirow{2}*{\shortstack{Random+KD\\$k$=196}} & Acc & 62.03 & 62.07 \\ 
        ~ & Computation & 75\% & 75\% \\
        
        \midrule
        \multirow{2}{*}{\shortstack{UNIXKD\\$k$=196}} & Acc & \textbf{62.92} & \textbf{63.64} \\
        ~ & Computation & 78.43\% & 78.92\% \\
        \bottomrule
    \end{tabular}
    }\label{tab:a-tiny}
    \vspace{-5pt}
\end{table}

\begin{table*}[t]
    \centering
    \caption{Results on cross-architecture pairs: CIFAR100 Top-1 accuracy and computation cost. \textbf{Bold} font denotes the result that outperforms KD+label. Our method averagely surpasses KD+label by 1.10\% with 79.44\% computation cost. Average on 4 runs.}
    \vspace{5pt}
    \small{
    \begin{tabular}{cc|cccccc}
        \toprule
        \multicolumn{2}{c|}{Teacher} & VGG13       & ResNet50      & ResNet50  & resnet32$\times$4 & resnet32$\times$4 & WRN-40-2  \\
        \multicolumn{2}{c|}{Student} & MobileNetV2 & MobileNetV2   & VGG8      & ShuffleV1         & ShuffleV2         & ShuffleV1 \\
        
        \midrule
        \multicolumn{2}{c|}{Teacher Acc} & 75.38 & 78.86 & 78.86 & 79.58 & 79.58 & 76.46 \\
        \multicolumn{2}{c|}{Student Acc} & 65.67 & 65.67 & 70.68 & 71.46 & 72.64 & 71.46 \\
        
        \midrule
        \multicolumn{2}{c|}{$F_t / F_s$} & 38.17 & 174.00 & 13.56 & 27.16 & 23.49 & 8.22\\
        
        \midrule
        \multirow{2}*{KD+label} & Acc & 68.32 & 68.24 & 73.50 & 74.07 & 74.93 & 76.00\\ 
        ~ & Computation & 100\% & 100\% & 100\% & 100\% & 100\% & 100\%\\
        
        \midrule
        \multirow{2}{*}{\shortstack{UNIXKD+label\\$k$=48}} & Acc & \textbf{68.78} & \textbf{68.96} & \textbf{73.88} & \textbf{76.28} & \textbf{76.69} & \textbf{77.04}\\
        ~ & Computation & 77.49\% & 75.57\% & 81.43\% & 78.43\% & 78.92\% & 84.79\%\\
        
        \bottomrule
    \end{tabular}
    }
    \label{tab:a-diff_arch}
\end{table*}

\begin{table*}[t]
    \centering
    \caption{Results on similar-architecture pairs: CIFAR100 Top-1 accuracy and computation cost. \textbf{Bold} font denotes the result that outperforms KD+label. Our method averagely surpasses KD+label by 0.53\% with 92.40\% computation cost. Average on 4 runs.}
    \vspace{5pt}
    \small{
    \begin{tabular}{cc|cccccc}
        \toprule
        \multicolumn{2}{c|}{Teacher} & WRN-40-2 & WRN-40-2 & resnet56 & resnet32$\times$4 & VGG13 & VGG16\\
        \multicolumn{2}{c|}{Student} & WRN-16-2 & WRN-40-1 & resnet20 & resnet8$\times$4 & VGG8 & VGG8\\
        
        \midrule
        \multicolumn{2}{c|}{Teacher Acc} & 76.20 & 76.20 & 73.10 & 79.58 & 74.73 & 74.76 \\
        \multicolumn{2}{c|}{Student Acc} & 73.45 & 72.02 & 69.38 & 73.22 & 70.98 & 70.98 \\
        
        \midrule
        \multicolumn{2}{c|}{$F_t / F_s$} & 3.25 & 3.93 & 3.06 & 6.12 & 2.97 & 4.14\\
        
        \midrule
        \multirow{2}*{KD+label} & Acc & 75.18 & 74.10 & 71.06 & 73.21 & 73.32 & 72.22\\ 
        ~ & Computation & 100\% & 100\% & 100\% & 100\% & 100\% & 100\%\\
        
        \midrule
        \multirow{2}{*}{\shortstack{UNIXKD+label\\$k$=48}} & Acc & \textbf{75.48} & \textbf{74.53} & 70.72 & \textbf{74.78} & \textbf{73.72} & \textbf{73.04} \\
        ~ & Computation & 94.06\% & 91.88\% & 94.76\% & 87.32\% & 95.10\% & 91.29\%\\
        
        \bottomrule
    \end{tabular}
    }
    \label{tab:a-same_arch}
    \vspace{-5pt}
\end{table*}

\subsection{Data Augmentation.}
For CIFAR100, we sequentially apply random crop, random horizontal flip and normalization. The crop size is 32$\times$32. 
For TinyImageNet, we adopt the same operations as CIFAR100, except that the crop size is 64$\times$64.
For ImageNet, we follow the official implementation of PyTorch~\cite{pytorch}.

\subsection{Training Hyperparameters}

\vspace{3pt}
\noindent\textbf{CIFAR100.}
The temperature of KD is set as 4 (same for TinyImageNet and ImageNet).
We train all the student models for 240 epochs. 
The initial learning rate of ShuffleV1, ShuffleV2 and MobileNetV2 is 0.01. 
For other architectures, the initial learning rate is 0.05.
The learning rate is decayed by a factor of 10, respectively, at 150, 180 and 210 epochs.
The $\alpha$ of beta distribution is $1.0$.
We run each experiment on a TITAN-X-Pascal GPU with a batch size of 64. An SGD optimizer with a $5\times10^{-4}$weight decay and 0.9 momentum is adopted.

When we include hard label loss into the learning objective, we use the common combination coefficient in KD:
\begin{equation}
    L = 0.1\cdot L_{ce} + 0.9\cdot L_{kd},
\end{equation}
where $L_{ce}$ is the cross entropy loss with hard label.

When we combine KD with AT~\cite{AT} and SP~\cite{simi}, we keep the loss weight of $L_{kd}$=1 and set loss weights of AT and SP to be $1,000$ and $3,000$, respectively.

\vspace{3pt}
\noindent\textbf{TinyImageNet.}
We train all the student models for 100 epochs.
The initial learning rate is 0.1 for all the architectures.
It is decayed by a factor of 10, respectively, at 40, 70 and 90 epochs.
The $\alpha$ of beta distribution is $0.2$.
We run each experiment on four parallel TITAN-X-Pascal GPU with a total batch size of 256. An SGD optimizer with a $5\times10^{-4}$weight decay and 0.9 momentum is adopted.

\vspace{3pt}
\noindent\textbf{ImageNet.}
We train all the student models for 90 epochs.
The initial learning rate is 0.1 and is decayed by a factor of 10, respectively, at 30 and 60 epochs.
The $\alpha$ of beta distribution is $0.2$.
We run each experiment on eight parallel TITAN-X-Pascal GPU with a total batch size of 256. An SGD optimizer with a $1\times10^{-4}$weight decay and 0.9 momentum is adopted.
When we combine KD with hard label loss, the learning objective is:
\begin{equation}
    L = L_{ce} + 0.9\cdot L_{kd},
\end{equation}
where $L_{ce}$ is the cross entropy loss with hard label.

\end{appendix}

\end{document}